\documentclass[conference]{IEEEtran}
\IEEEoverridecommandlockouts
\usepackage{cite}
\usepackage{amsmath,amssymb,amsfonts}
\usepackage{algorithmic}
\usepackage{graphicx}
\usepackage{textcomp}
\usepackage{xcolor}
\def\BibTeX{{\rm B\kern-.05em{\sc i\kern-.025em b}\kern-.08em
    T\kern-.1667em\lower.7ex\hbox{E}\kern-.125emX}}
\usepackage{acro}
\usepackage{mathtools}
\usepackage{nicefrac}
\usepackage{siunitx}
\usepackage{tikz}
\usepackage{tikz-3dplot}
\usetikzlibrary{plotmarks}
\usetikzlibrary{patterns.meta}
\usetikzlibrary{svg.path}
\usepackage{xcolor}
\usepackage{scalerel}

\usepackage[utf8]{inputenc}
\usepackage{pgfplots}
\DeclareUnicodeCharacter{2212}{−}
\usepgfplotslibrary{groupplots,dateplot}
\usetikzlibrary{patterns,shapes.arrows}
\pgfplotsset{compat=newest}
\usepackage{multirow}
\usepackage{hyperref}

\DeclareAcronym{MLTP}{
	short=MLTP,
	long=minimum lap time problem
}

\DeclareAcronym{OCP}{
	short=OCP,
	long=optimal control problem
}

\DeclareAcronym{QSS}{
	short=QSS,
	long=quasi-steady-state
}

\DeclareAcronym{3D}{
	short=3D,
	long=three-dimensional
}

\DeclareAcronym{2D}{
	short=2D,
	long=two-dimensional
}

\DeclareAcronym{MPC}{
	short=MPC,
	long=model predictive control
}

\DeclareAcronym{MPCB}{
	short=MPCB,
	long=Mount Panorama Circuit in Bathurst
}

\DeclareAcronym{LVMS}{
	short=LVMS,
	long=Las Vegas Motor Speedway
}

\DeclareAcronym{NLP}{
	short=NLP,
	long=nonlinear program
}

\DeclareAcronym{CoM}{
	short=CoM,
	long=center of mass,
}

\DeclareAcronym{SQP}{
	short=SQP,
	long=sequential quadratic programming,
}

\definecolor{TUMBlue}{HTML}{0065BD}
\definecolor{TUMBlack}{HTML}{000000}
\definecolor{TUMWhite}{HTML}{ffffff}
\definecolor{TUMBlueLight}{HTML}{005293}
\definecolor{TUMBlueDark}{HTML}{003359}
\definecolor{TUMGrayDark}{HTML}{333333}
\definecolor{TUMGray}{HTML}{808080}
\definecolor{TUMGrayLight}{HTML}{CCCCCC}
\definecolor{TUMAccGray}{HTML}{DAD7CB}
\definecolor{TUMAccGreen}{HTML}{A2AD00}
\definecolor{TUMAccOrange}{HTML}{E37222}
\definecolor{TUMAccBlueLight}{HTML}{98C6EA}
\definecolor{TUMAccBlueDark}{HTML}{64A0C8}

\newcommand*{\circled}[2][]{\tikz[baseline=(C.base)]{
		\node[inner sep=0pt] (C) {\footnotesize \vphantom{1g}#2};
		\node[draw, circle, inner sep=1.3pt, yshift=1pt] 
		at (C.center) {\vphantom{1g}};}}
	
\pgfkeys{/pgf/number format/.cd,1000 sep={}}
\pgfplotsset{
	compat=1.11,
	legend image code/.code={
		\draw[mark repeat=2,mark phase=2,line width=2 pt]
		plot coordinates {
			(0cm,0cm)
			(0.1cm,0cm)        %% default is (0.3cm,0cm)
			(0.29cm,0cm)         %% default is (0.6cm,0cm)
		};%
	}
}

\usepackage{framed}
\begin{document}
	
\begin{minipage}{\textwidth}
	\centering
	\begin{framed}
		\copyright~2023 IEEE. Personal use of this material is permitted. Permission from IEEE must be obtained for all other uses, in any current or future media, including reprinting/republishing this material for advertising or promotional purposes, creating new collective works, for resale or redistribution to servers or lists, or reuse of any copyrighted component of this work in other works.
	\end{framed}
\end{minipage}

\title{Online Time-Optimal Trajectory Planning on Three-Dimensional Race Tracks}

\author{\IEEEauthorblockN{Matthias Rowold\IEEEauthorrefmark{1}, Levent Ögretmen, Ulf Kasolowsky, and Boris Lohmann}\thanks{All authors are with the Chair of Automatic Control, Department of Mechanical Engineering, TUM School of Engineering and Design, Technical University of Munich, 85748 Garching, Germany.\newline\IEEEauthorrefmark{1} Corresponding author: \href{mailto:matthias.rowold@tum.de}{matthias.rowold@tum.de}}}

\maketitle

\begin{abstract}
We propose an online planning approach for racing that generates the time-optimal trajectory for the upcoming track section. The resulting trajectory takes the current vehicle state, effects caused by \acl{3D} track geometries, and speed limits dictated by the race rules into account. In each planning step, an \acl{OCP} is solved, making a \acl{QSS} assumption with a point mass model constrained by gg-diagrams. For its online applicability, we propose an efficient representation of the gg-diagrams and identify negligible terms to reduce the computational effort. We demonstrate that the online planning approach can reproduce the lap times of an offline-generated racing line during single vehicle racing. Moreover, it finds a new time-optimal solution when a deviation from the original racing line is necessary, e.g., during an overtaking maneuver. Motivated by the application in a rule-based race, we also consider the scenario of a speed limit lower than the current vehicle velocity. We introduce an initializable slack variable to generate feasible trajectories despite the constraint violation while reducing the velocity to comply with the rules.
\end{abstract}


\section{Introduction}
A racing line is an optimal path or trajectory around a race track. In the absence of additional objectives such as minimization of fuel consumption or tire wear, this is usually the time-optimal trajectory obtained by solving a \ac{MLTP}. Recent approaches to trajectory planning for autonomous race cars use an offline-generated racing line as a reference for a local planning algorithm \cite{Stahl2019, Raji2022, Rowold2022, Ogretmen2022}. However, the racing line is only time-optimal as long as the vehicle follows it spatially and temporally. If a deviation from the racing line is necessary, e.g., due to an evasion maneuver, the offline-calculated racing line loses its optimality. Furthermore, low-speed situations (e.g., race start, speed limits) represent scenarios in which, in terms of time-optimality, a different trajectory closer to the shortest path outperforms a classic apex-to-apex racing line. These scenarios motivate to regularly update the racing line during the race according to the current vehicle state and the applicable speed limit.

\begin{figure}
	\centering
	\includegraphics[width=8cm,trim={2cm 0 0 3cm},clip]{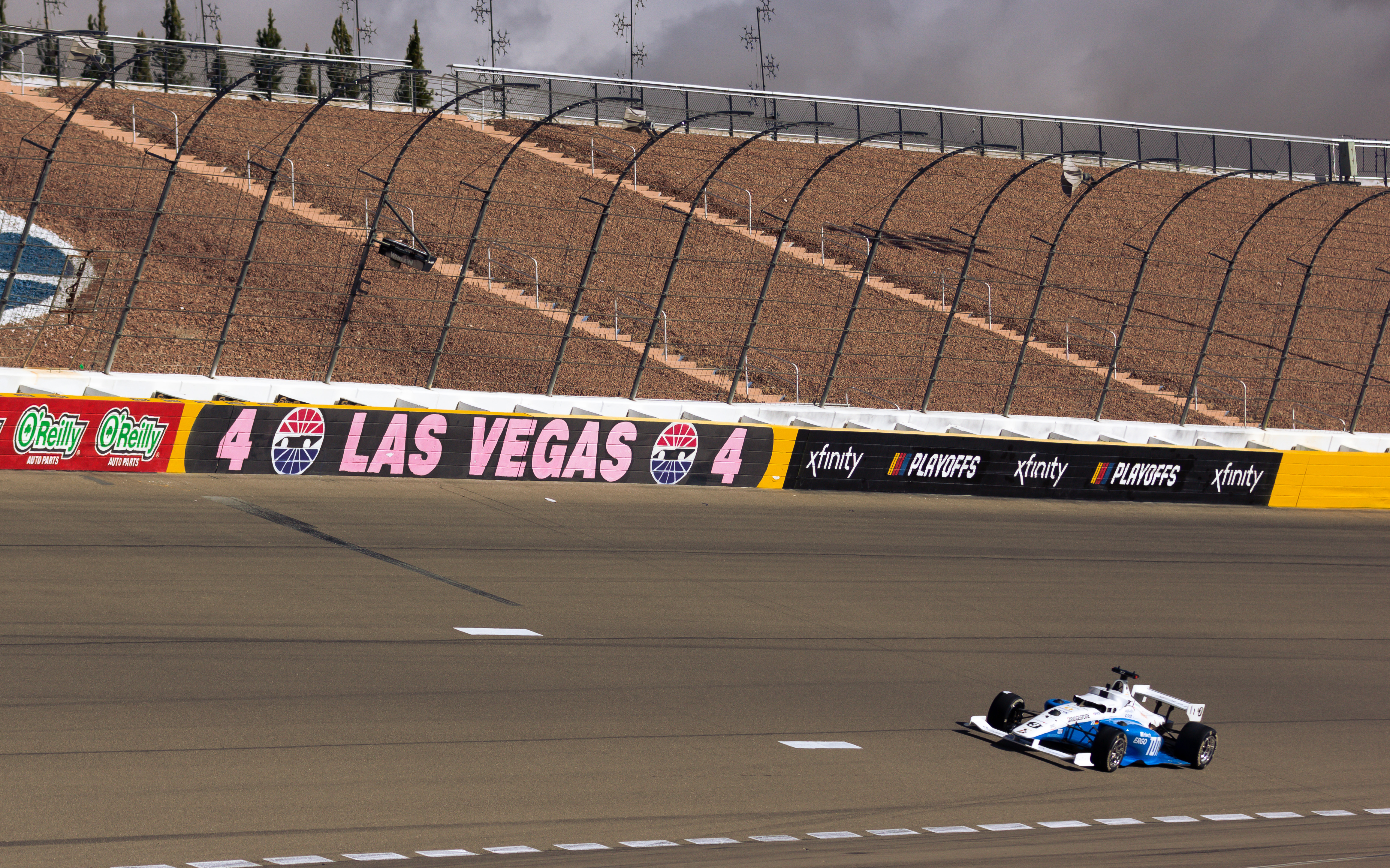}
	\caption{Team TUM Autonomous Motorsport's racing car, Dallara AV-21, in a banked turn on the Las Vegas Motor Speedway.}
	\label{fig:banked_turn}
\end{figure}

We propose an online planning approach for racing that generates the time-optimal trajectory for the upcoming track section. The result of this approach is called the local racing line. Compared to the closed racing line around the race track, which will be referred to as the global racing line hereafter, the local racing line has a limited spatial horizon. In addition to the current vehicle state and the speed limit, our approach considers \acf{3D} race track geometries as they appear in recent events with full-scale prototypes, as shown in Figure~\ref{fig:banked_turn} \cite{Wischnewski2023, Raji2022}. We show that the \ac{3D} geometries significantly influence the achievable lap times and should therefore be considered in the online planning approach. We demonstrate the ability to reproduce the lap time of the global racing line during single vehicle racing and to update the local racing line when a deviation from the global racing line is necessary or a speed limit is announced.

\subsection{Related Work}
The survey in \cite{Betz2022} provides an overview of perception, planning, and control algorithms for autonomous racing. The authors consider the task of global planning to be the generation of a racing line around the race track. The start and end states of the racing line are the same to achieve the minimum lap time for a flying start. To solve the \ac{MLTP}, many authors have formulated an \acf{OCP} with a detailed vehicle model \cite{Casanova2000, Rucco2015, Limebeer2015, Christ2021}. However, the resulting \ac{OCP} can be large, depending on the track length, discretization, and the number of states, resulting in computation times that are not applicable online. \Acf{QSS} approaches use simpler point mass models and thus achieve shorter computation times \cite{Veneri2020, Lovato2022}. They assume successive stationary states, each corresponding to a stationary circular drive for a specific radius, to form a trajectory. The states must lie within pre-calculated velocity-dependent gg-diagrams that constrain the longitudinal and lateral accelerations.

Some approaches also perform the global racing line generation for \ac{3D} race tracks. Based on the \ac{3D} track representation with laterally flat ribbons in \cite{Perantoni2015}, the \ac{QSS} approach in \cite{Lovato2022} generates the racing line for tracks with varying slope and banking angles. An \ac{OCP} using a double-track model is presented in \cite{Limebeer2015}. Lovato et al. \cite{Lovato2021} extend the \ac{3D} representation with laterally curved ribbons that allow modeling progressively banked turns.

The above approaches generate the global racing line which can be used as a fixed reference for a subsequent local planning algorithm. The following two approaches perform an online re-planning for a limited horizon and thus counteract the mentioned problems caused by a deviation from the racing line. Herrmann et al. \cite{Herrmann2021} optimize the velocity profile starting from the current vehicle state, but they assume a given path with a length of \SI{300}{\meter} and do not consider speed limits. Gundlach and Konigorski \cite{Gundlach2019} formulate an \ac{OCP} to find the time-optimal path and velocity profile. Their variable discretization allows for a planning horizon of up to \SI{480}{\meter} and they foster the existence of a feasible follow-up trajectory by penalizing the deviation of the end state from the global racing line. However, in low-speed situations or during applicable speed limits, the global racing line does not represent the time-optimal path, so returning to it may be undesirable. Both online planning approaches do not consider \ac{3D} track geometries or provide a method to handle speed limits lower than the current velocity.

\subsection{Contributions}
The following contributions relate to the generation of a local racing line and the associated challenges for the online application:
\begin{itemize}
	\item We present an online planning approach based on an \ac{OCP} with a \ac{QSS} assumption to generate the local racing line on a \ac{3D} race track considering the current state and the speed limit. To keep the \ac{OCP} solvable in situations where the current vehicle speed is greater than a new speed limit, we introduce an initializable slack variable.
	\item We develop an efficient representation of \ac{3D} gg-diagrams that is advantageous for convergence speed and thus suitable for an online application. Despite being an underapproximation of the true gg-diagrams, it allows reasonable lap times.
	\item We provide an analysis for negligible terms and analyze the updated local racing line in cases where the time-optimal solution differs from the global racing line.
\end{itemize}

\subsection{Notation and Conventions}
We use right-handed coordinate systems and define our inertial frame of reference $\mathcal{I}$ with the x, y, and z-axes pointing east, north, and up, respectively. $\mathbf{R}_\mathrm{x}(\gamma)$, $\mathbf{R}_\mathrm{y}(\gamma)$, $\mathbf{R}_\mathrm{z}(\gamma)$ denote the rotation matrices for an angle $\gamma$ around the x, y, and \mbox{z-axis}. $\sin(\gamma)$ and $\cos(\gamma)$ are sometimes abbreviated by $s_\gamma$ and $c_\gamma$. A subscript before a vector (bold) indicates the coordinate frame in which the vector is expressed. E.g., ${}_\mathcal{I}\mathbf{r}$ describes $\mathbf{r}$ in the inertial frame. A postscript after a vector indicates the point to which the vector refers. E.g., ${}_\mathcal{I}\mathbf{v}_\mathrm{CoM}$ expresses the velocity of the vehicle's \ac{CoM} in the intertial frame $\mathcal{I}$. Derivatives with respect to the arc length $s$ are noted as $\frac{d\square}{ds} = \square^\prime$ and with respect to time $t$ as $\frac{d\square}{dt} = \dot{\square}$.

\section{Methodology}
In Sections \ref{sec:track_representation} and \ref{sec:vehicle_model}, we summarize the needed relations for the \ac{3D} track representation and vehicle model as they are presented in \cite{Veneri2020, Perantoni2015, Lovato2021, Lovato2022}. Section~\ref{sec:constraints} describes the efficient representation of gg-diagrams, and Section~\ref{sec:time_optimal_ocp} formulates the time-optimal \ac{OCP} for generating global and local racing lines.

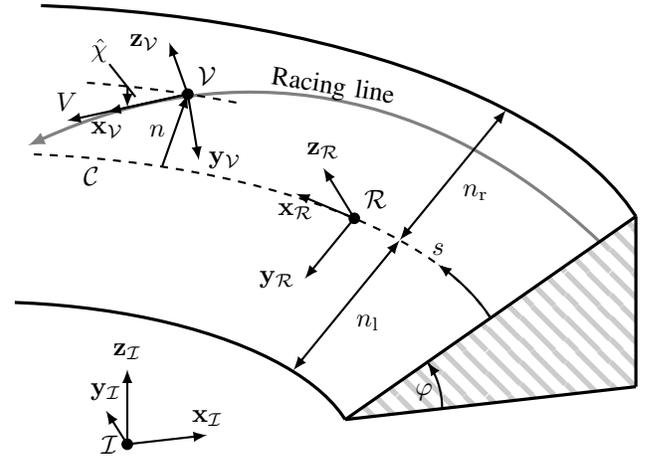
\begin{figure}[]
	\centering
	% !TeX root = ../main.tex
 \tdplotsetmaincoords{65}{-15}
 \def\width{4}
 \def\height{2.5}
 \def\banking{-atan(\height/\width)}
 \def\radius{5}
 \def\sec{70}
 \def\Rsec{30}
 \def\Vsec{55}
 \def\Axlength{1.1}
 \def\chihat{25.0}
\begin{tikzpicture}[tdplot_main_coords]
	% trajectory
	\draw[color=TUMGray, very thick, -latex] (7*\width/8, 0, {7*\height/8}) to [bend right=34.7] node[above=-3.5,rotate=-9,text=TUMBlack]{Racing line} ({(\radius + \width/2)*cos(\sec)-\radius}, {(\radius + \width/2)*sin(\sec)}, {\height*0.55});
	
	% triangle
	\path[pattern={Lines[angle=-45,distance={6},line width=2]},pattern color=TUMGrayLight] (0, 0, 0) -- (\width, 0, 0) --  (\width, 0, \height);
	\draw[very thick] (0 ,0, 0) -- (\width, 0, 0);
	\draw[very thick] (0 ,0, 0) -- (\width, 0, \height);
	\draw[very thick] (\width, 0, 0) -- (\width, 0, \height);
	
	% banking angle
	\draw[thick, domain=0:{-\banking}, -latex]  plot ({\width/3*cos(\x)}, 0, {\width/3*sin(\x)});
	\node[left] at ({\width/3}, 0, {\width/3*sin(-\banking/3)}) {$\varphi$};
	
	% track bounds
	\draw[very thick, domain=0:\sec]  plot ({\radius*cos(\x)-\radius}, {\radius*sin(\x)}, 0);
	\draw[very thick, domain=0:{\sec}]  plot ({(\radius + \width)*cos(\x)-\radius}, {(\radius + \width)*sin(\x)}, \height);
	
	% center line
	\draw[dashed, thick, domain=0:{\sec}]  plot  ({(\radius + \width/2)*cos(\x)-\radius}, {(\radius + \width/2)*sin(\x)}, {\height/2});
	\node[below] at ({(\radius + \width/2)*cos(\sec*(9/10))-\radius}, {(\radius + \width/2)*sin(\sec*(9/10))}, {\height/2}){$\mathcal{C}$};
	%s
	\draw[thick, -latex, domain=0:{2*\Rsec/4}]  plot ({(\radius + \width/2)*cos(\x)-\radius}, {(\radius + \width/2)*sin(\x)}, {\height/2});
	\node[above] at ({(\radius + \width/2)*cos(4*\Rsec/8)-\radius}, {(\radius + \width/2)*sin(4*\Rsec/8)}, {\height/2}) {$s$};

	% inertial frame
	\node at (-3 ,0, 0) {\pgfuseplotmark{*}};
	\draw[thick, -latex] (-3 ,0, 0)node[left]{$\mathcal{I}$} -- +(\Axlength, 0, 0)node[above]{$\mathbf{x}_\mathcal{I}$};
	\draw[thick, -latex] (-3 ,0, 0) -- +(0, \Axlength, 0)node[above]{$\mathbf{y}_\mathcal{I}$};
	\draw[thick, -latex] (-3, 0, 0) -- +(0, 0, \Axlength)node[above]{$\mathbf{z}_\mathcal{I}$};
	
	% road frame
	\def\xR{(\radius + \width/2)*cos(\Rsec)-\radius}
	\def\yR{(\radius + \width/2)*sin(\Rsec)}
	\node at ({\xR} , {\yR}, {\height/2}) {\pgfuseplotmark{*}};
	\draw[thick, -latex] ({\xR} , {\yR}, {\height/2})node[above right]{$\mathcal{R}$} -- +({cos(90 + \Rsec)*\Axlength}, {sin(90 + \Rsec)*\Axlength}, 0)node[below]{$\mathbf{x}_\mathcal{R}$};
	\draw[thick, -latex] ({\xR} , {\yR}, {\height/2}) -- +({-sin(90 + \Rsec)*cos(\banking)*\Axlength}, {cos(90 + \Rsec)*cos(\banking)*\Axlength}, {sin(\banking)*\Axlength})node[left]{$\mathbf{y}_\mathcal{R}$};
	\draw[thick, -latex] ({\xR} , {\yR}, {\height/2}) -- +({sin(90+\Rsec)*sin(\banking)*\Axlength}, {-cos(90+\Rsec)*sin(\banking)*\Axlength}, {cos(\banking)*\Axlength})node[above]{$\mathbf{z}_\mathcal{R}$};
	
	% velocity frame
	\def\xV{(\radius + 3*\width/4)*cos(\Vsec)-\radius}
	\def\yV{(\radius + 3*\width/4)*sin(\Vsec)}
	\node at ({\xV} , {\yV}, {3*\height/4}) {\pgfuseplotmark{*}};
	\draw[thick, -latex] ({\xV} , {\yV}, {3*\height/4})node[above right]{$\mathcal{V}$} -- +({(-sin(\chihat)*sin(90 + \Vsec)*cos(\banking)+cos(\chihat)*cos(90+\Vsec))*\Axlength}, {(sin(\chihat)*cos(\banking)*cos(90+\Vsec)+sin(90+\Vsec)*cos(\chihat))*\Axlength}, {(sin(\chihat)*sin(\banking)*\Axlength)})node[below]{$\mathbf{x}_\mathcal{V}$};
	\draw[thick, -latex] ({\xV} , {\yV}, {3*\height/4}) -- +({(-sin(\chihat)*cos(90+\Vsec)-sin(90+\Vsec)*cos(\chihat)*cos(\banking))*\Axlength}, {(-sin(\chihat)*sin(90+\Vsec)+cos(\chihat)*cos(\banking)*cos(90+\Vsec))*\Axlength}, {(sin(\banking)*cos(\chihat))*\Axlength})node[right]{$\mathbf{y}_\mathcal{V}$};
	\draw[thick, -latex] ({\xV} , {\yV}, {3*\height/4}) -- +({sin(90+\Vsec)*sin(\banking)*\Axlength}, {-cos(90+\Rsec)*sin(\banking)*\Axlength}, {cos(\banking)*\Axlength})node[left]{$\mathbf{z}_\mathcal{V}$};
	
	%n
	\draw[thick, -latex] ({(\radius + \width/2)*cos(\Vsec)-\radius}, {(\radius + \width/2)*sin(\Vsec)}, {\height/2}) --node[left]{$n$} ({\xV} , {\yV}, {3*\height/4});
	%V
	\draw[thick, -latex] ({\xV} , {\yV}, {3*\height/4}) -- +({(-sin(\chihat)*sin(90 + \Vsec)*cos(\banking)+cos(\chihat)*cos(90+\Vsec))*\Axlength*1.5}, {(sin(\chihat)*cos(\banking)*cos(90+\Vsec)+sin(90+\Vsec)*cos(\chihat))*\Axlength*1.5}, {(sin(\chihat)*sin(\banking)*\Axlength*1.5)})node[above]{$V$};
	
	%chihat
	\draw[dashed, thick, domain={\Vsec-5}:{\Vsec+10}]  plot ({(\radius + 3*\width/4)*cos(\x)-\radius}, {(\radius + 3*\width/4)*sin(\x)}, {3*\height/4});
	\draw[thick, -latex] ({(\radius + 3*\width/4)*cos(\Vsec+6)-\radius}, {(\radius + 3*\width/4)*sin(\Vsec+6)}, {3*\height/4}) --node[right=6, below=0](nodetmp){} ({\xV + (-sin(\chihat)*sin(90 + \Vsec)*cos(\banking)+cos(\chihat)*cos(90+\Vsec))*\Axlength*0.75} , {\yV + (sin(\chihat)*cos(\banking)*cos(90+\Vsec)+sin(90+\Vsec)*cos(\chihat))*\Axlength*0.75}, {3*\height/4 + (sin(\chihat)*sin(\banking)*\Axlength*0.75)});
	\draw[thick] (nodetmp) -- +(-0.2, 1, 0.2)node[above left=-3]{$\hat{\chi}$};

	%nl
	\draw[thick, latex-latex] ({(\radius + \width/2)*cos(6*\Rsec/8)-\radius}, {(\radius + \width/2)*sin(6*\Rsec/8)}, {\height/2}) --node[below right]{$n_\mathrm{l}$} +({-sin(90 + \Rsec)*cos(\banking)*(\width/2/cos(-\banking))}, {cos(90 + \Rsec)*cos(\banking)*(\width/2/cos(-\banking))}, {sin(\banking)*(\width/2/cos(-\banking))});
	%nr
	\draw[thick, latex-latex] ({(\radius + \width/2)*cos(6*\Rsec/8)-\radius}, {(\radius + \width/2)*sin(6*\Rsec/8)}, {\height/2}) --node[below right]{$n_\mathrm{r}$} +({-sin(90 + \Rsec)*cos(\banking)*(-\width/2/cos(-\banking))}, {cos(90 + \Rsec)*cos(\banking)*(-\width/2/cos(-\banking))}, {sin(\banking)*(-\width/2/cos(-\banking))});

\end{tikzpicture}
	\caption{Banked left turn without slope ($\mu=0$) and with a negative banking angle ($\varphi<0$): The vehicle is located at a negative lateral displacement ($n<0$) relative to the reference line (dashed line). The relative orientation of the total velocity vector in the road plane is positive ($\hat{\chi}>0$).}
	\label{fig:3d_track}
\end{figure}
\subsection{\ac{3D} Track Representation}
\label{sec:track_representation}
We follow the ribbon-based track representation in \cite{Perantoni2015} to describe laterally flat \ac{3D} tracks. Figure~\ref{fig:3d_track} illustrates the introduced coordinate systems for a banked left turn. The spine of the ribbon is a \ac{3D} curve and will be referred to as the reference line $\mathcal{C}$:
\begin{equation}
	\label{eq:ref_line}
	\mathcal{C} = \left\{{}_\mathcal{I}\mathbf{p}(s) = \left[p_\mathrm{x}(s),p_\mathrm{y}(s),p_\mathrm{z}(s)\right]^\top \in \mathbb{R} ^3, s \in\left[0, s_\mathrm{f}\right]\right\}\text{.}
\end{equation}
A point ${}_\mathcal{I}\mathbf{p}(s)$ on the reference line is parameterized by the arc length $s$, which reaches from $0$ to the track length $s_\mathrm{f}$ with ${}_\mathcal{I}\mathbf{p}(0) = {}_\mathcal{I}\mathbf{p}(s_\mathrm{f})$. The tangential vector of the reference line $\mathbf{p}^\prime(s)$ defines the x-axis $\mathbf{x}_\mathcal{R}(s)$ of the road frame $\mathcal{R}$. The y-axis $\mathbf{y}_\mathcal{R}(s)$ is given by the normal vector in the road plane pointing from the reference line towards the left border of the track. The z-axis completes the frame with $\mathbf{z}_\mathcal{R}(s) = \mathbf{x}_\mathcal{R}(s) \times \mathbf{y}_\mathcal{R}(s)$. It is perpendicular to the road plane and points upwards. With the road frame defined, the road surface $\mathcal{S}$ can be parameterized by
\begin{equation}
	\begin{split}
		\mathcal{S} = \left\{{}_\mathcal{I}\mathbf{r}(s, n)\right. &= {}_\mathcal{I}\mathbf{p}(s) + {}_\mathcal{I}\mathbf{y}_\mathcal{R}(s)n \in \mathbb{R}^3, \\[1pt]
		&\left. s \in \left[0, s_\mathrm{f}\right], n \in [n_\mathrm{l}(s), n_\mathrm{r}(s)]\right\}\text{.}
	\end{split}
\end{equation}
While $s$ describes the progress along the reference line, $n$ specifies the lateral displacement. The distances to the left and right track boundaries $n_\mathrm{l}(s)$ (positive) and $n_\mathrm{r}(s)$ (negative) at the respective $s$ constrain $n$. For simplicity, we omit the dependency on $s$ in the following equations.

The orientation of the road frame $\mathcal{R}$ with respect to the inertial frame $\mathcal{I}$ is expressed by zyx Euler angles, where $\theta$ is the rotation around the z-axis, $\mu$ the rotation around the resulting y-axis, and $\varphi$ the rotation around the resulting x-axis. In the remainder of this paper, we will denote $\theta$ as the orientation, $\mu$ as the slope, and $\varphi$ as the banking angle. The angular velocity $\mathbf{\Omega}_\mathcal{R}$ of the road frame with respect to the arc length, expressed in the road frame itself, is obtained by
\begin{equation}
	{}_\mathcal{R}\mathbf{\Omega}_\mathcal{R} = \begin{bmatrix}
		\Omega_\mathrm{x} \\
		\Omega_\mathrm{y} \\
		\Omega_\mathrm{z}
	\end{bmatrix} = \begin{bmatrix}
		1 & 0 & -s_\mu \\
		0 & c_\varphi & c_\mu s_\varphi \\
		0 & -s_\varphi & c_\mu c_\varphi
	\end{bmatrix}\begin{bmatrix}
		\varphi^\prime\\
		\mu^\prime\\
		\theta^\prime
	\end{bmatrix}\text{.}
\end{equation}
We solve an optimization problem, as described in \cite{Perantoni2015}, to smooth the resulting track data and ensure continuity at the start-finish line.

\subsection{Vehicle Model}
\label{sec:vehicle_model}
Following \cite{Veneri2020} and \cite{Lovato2022}, we model the vehicle as a point mass. As pointed out in \cite{Lovato2022}, a distinction must be made between the vehicle frame, whose x-axis aligns with the vehicles' longitudinal axis, and the velocity frame $\mathcal{V}$. The x-axis of $\mathcal{V}$ coincides with the total velocity vector in the road plane, as shown in Figure~\ref{fig:3d_track}, so there is no lateral velocity component in the direction of its y-axis. The difference between the two frames is a rotation around the z-axis by the slip angle $\beta$. With the vehicle's orientation relative to the reference line $\chi$, the relative orientation of $\mathcal{V}$ is $\hat{\chi} = \chi + \beta$. 

Given the total velocity $V$ of the point mass in the road plane and a lateral deviation from the reference line $n$, the velocity along the reference line is
\begin{equation}
	\label{eq:s_dot}
	\dot{s} = \frac{ds}{dt} = \frac{V\cos(\hat{\chi})}{1 - n \Omega_\mathrm{z}}\text{.}
\end{equation}
The angular velocity of the road frame with respect to time then is ${}_\mathcal{R}\boldsymbol{\omega}_\mathcal{R} = \begin{bmatrix}	\omega_\mathrm{x} & \omega_\mathrm{y} & \omega_\mathrm{z}\end{bmatrix}^\top = {}_\mathcal{R}\mathbf{\Omega}_\mathcal{R}\dot{s}$. 
With a varying relative orientation $\hat{\chi}$, the angular velocity of the velocity frame $\mathcal{V}$ with respect to time is
\begin{equation}
	\label{eq:omega_in_vel}
	\begin{split}
		{}_\mathcal{V}\boldsymbol{\omega}_\mathcal{V} = \begin{bmatrix}
			\hat{\omega}_\mathrm{x} \\
			\hat{\omega}_\mathrm{y} \\
			\hat{\omega}_\mathrm{z} \\
		\end{bmatrix} &= \mathbf{R}_\mathrm{z}^\top(\hat{\chi}){}_\mathcal{R}\boldsymbol{\omega}_\mathcal{R} +\begin{bmatrix}
			0 \\ 0 \\ \dot{\hat{\chi}}
		\end{bmatrix} = \begin{bmatrix}
			\left(\Omega_\mathrm{x}c_{\hat{\chi}} + \Omega_\mathrm{y}s_{\hat{\chi}}\right)\dot{s} \\
			\left(\Omega_\mathrm{y}c_{\hat{\chi}} - \Omega_\mathrm{x}s_{\hat{\chi}}\right)\dot{s} \\
			\Omega_\mathrm{z}\dot{s} + \dot{\hat{\chi}}
		\end{bmatrix}\hspace{-0.1cm}\text{.}
	\end{split}
\end{equation}
With the vertical velocity $w=n\omega_\mathrm{x}$ and its derivative ${\dot{w}=\dot{n}\omega_\mathrm{x} + n\dot{\omega}_\mathrm{x}}$, the acceleration of the velocity frame $\mathcal{V}$ is given by
\begin{equation}
	\label{eq:acc_in_vel}
	{}_\mathcal{V}\mathbf{a}_\mathcal{V} = \begin{bmatrix}
		\hat{a}_\mathrm{x} \\
		\hat{a}_\mathrm{y} \\
		\hat{a}_\mathrm{z} \\
	\end{bmatrix} = \begin{bmatrix}
		\dot{V}\\
		0\\
		\dot{w}
	\end{bmatrix} + {}_\mathcal{V}\boldsymbol{\omega}_\mathcal{V} \times \begin{bmatrix}
		V\\
		0\\
		w
	\end{bmatrix} = \begin{bmatrix}
		\dot{V} + \hat{\omega}_\mathrm{y}w \\
		\hat{\omega}_\mathrm{z}V - \hat{\omega}_\mathrm{x}w \\
		\dot{w} - \hat{\omega}_\mathrm{y}V 
	\end{bmatrix}\text{.}
\end{equation}
We choose the state of the point mass moving with the velocity frame to
\begin{equation}
	\label{eq:state_vector}
	\mathbf{x} = \begin{bmatrix}	V & n & \hat{\chi} & \hat{a}_\mathrm{x} & \hat{a}_\mathrm{y} \end{bmatrix}^\top\text{.}
\end{equation}
With $\dot{n}=V\sin(\hat{\chi})$, the first two rows of \eqref{eq:acc_in_vel}, the last row of \eqref{eq:omega_in_vel}, and the longitudinal and lateral jerks as inputs ${\mathbf{u} = \begin{bmatrix} \hat{j}_\mathrm{x} & \hat{j}_\mathrm{y} \end{bmatrix}^\top}$, the dynamics are described by
\begin{equation}
	\label{eq:dynamics}
	\dot{\mathbf{x}} = \frac{d\mathbf{x}}{dt}=\mathbf{f}(\mathbf{x}, \mathbf{u}) = \begin{bmatrix}
		\hat{a}_\mathrm{x} - w \hat{\omega}_\mathrm{y}\\
		V\sin(\hat{\chi}) \\
		\frac{\hat{a}_\mathrm{y} + w\hat{\omega}_\mathrm{x}}{V}-\Omega_\mathrm{z}\dot{s}\\
		\hat{j}_\mathrm{x}\\
		\hat{j}_\mathrm{y}
	\end{bmatrix}\text{.}
\end{equation}
The longitudinal and lateral accelerations $\hat{a}_\mathrm{x}$ and $\hat{a}_\mathrm{y}$ of the point mass are the decisive variables that must be constrained in order to represent the actual, more complex vehicle. This is achieved with gg-diagrams that specify the admissible combinations of longitudinal and lateral accelerations. The generation of such gg-diagrams is usually computationally intensive but can be carried out offline. Once generated, they are used online in the \ac{OCP} with the low-dimensional point mass model, which achieves lower computation times than a higher-dimensional vehicle model.

The gg-diagrams for constraining the accelerations are based on a \ac{QSS} assumption, which we demonstrate in Section~\ref{sec:constraints}. The resulting trajectory is a sequence of points, each describing a state of stationary circular driving at a specific radius, neglecting transient effects. However, as long as the change between two successive states is small, the trajectory can be considered feasible. As done in Section~\ref{sec:time_optimal_ocp}, this can be achieved by regularizing the jerk.

\subsection{Acceleration Constraints}
\label{sec:constraints}
Polytopic state constraints limit the velocity $V$, the relative orientation $\hat{\chi}$, and the lateral deviation $n$. They are introduced in detail in Section~\ref{sec:time_optimal_ocp}. For the \ac{2D} case, the constraints on the accelerations depend on the velocity $V$ and are non-linearly coupled. They are described by gg-diagrams obtained from measured data or numerical simulations. Lovato and Massaro \cite{Lovato2022} extend the velocity-dependent gg-diagrams to introduce the dependency on the vertical acceleration $\tilde{g}$, which can be different from $g=\SI{9.81}{\meter\per\second\squared}$ on \ac{3D} tracks. 
The shape and size of a gg-diagram are now influenced by two parameters, namely $V$ and $\tilde{g}$. $\tilde{g}$ accounts for the gravity and effects due to the \ac{3D} motion.

\begin{figure}
	\small
	\centering
	\def\axiswidth{7.0cm}
	\def\axisheight{4cm}
	\input{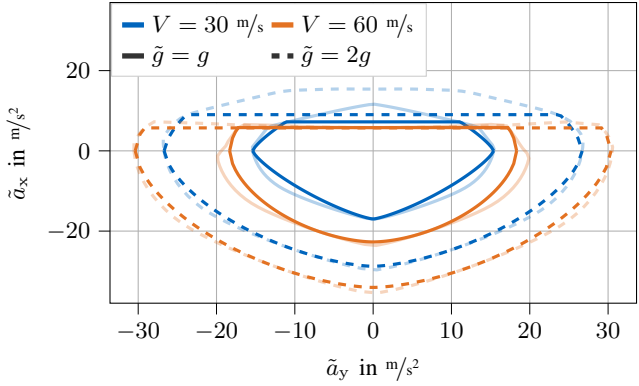}
	\caption{Representation of gg-diagrams: The transparent lines are obtained by the approach in \cite{Veneri2020}. The opaque lines show the underapproximating shapes according to \eqref{eq:gg_ax_max}--\eqref{eq:gg_combined}. The optimized values for $p$ are $p(30, g)=1.2$, $p(30, 2g)=1.6$, $p(60, g)=1.7$, and $p(60, 2g)=1.6$.}
	\label{fig:gg_diagram}
\end{figure}

To generate the gg-diagrams, we follow the method in \cite{Veneri2020} using a double-track model with longitudinal and lateral load transfer and a simplified Pacejka tire model. For the \ac{3D} case, the accelerations acting on the \ac{CoM} are the apparent accelerations $\tilde{a}_\mathrm{x}$, $\tilde{a}_\mathrm{y}$, and $\tilde{g}$, which are here expressed in the vehicle frame. A \ac{NLP} is formulated to find the stationary state that maximizes the combined acceleration $\textstyle\sqrt{\tilde{a}_\mathrm{x}^2+ \tilde{a}_\mathrm{y}^2}$ for a given combination of the total velocity $V$, apparent vertical acceleration $\tilde{g}$, and angle ${\alpha=\arctan(\nicefrac{\tilde{a}_\mathrm{y}}{\tilde{a}_\mathrm{x}})}$. The stationary state can be transformed into the velocity frame $\mathcal{V}$ by rotating around the z-axis by $\beta$ so that we consider $\tilde{a}_\mathrm{x}$, $\tilde{a}_\mathrm{y}$, and $\tilde{g}$ to be expressed in $\mathcal{V}$ from here on. The resulting gg-diagrams are visualized by the transparent lines in Figure~\ref{fig:gg_diagram} and show how $V$ and $\tilde{g}$ affect the acceleration limits. As the velocity increases, the maximum longitudinal acceleration decreases due to the increased drag force. At the same time, the maximum lateral acceleration increases due to the increased downforce. An increased vertical acceleration of $\tilde{g}=2g$, which can readily be achieved in banked turns on oval race tracks, induces an expansion of the gg-diagram, albeit to a lesser extent for positive longitudinal accelerations.

The polar coordinate parameterization in \cite{Lovato2022} with an adherence radius $\rho(V, \tilde{g}, \alpha)$ for $\alpha\in\left[-\pi, \pi\right)$ can represent the true shape of a gg-diagram for a given $V$ and $\tilde{g}$. However, this parameterization is disadvantageous regarding computation times when applied directly in the time-optimal \ac{OCP}. Therefore, we propose to use simpler diamond shapes that require only four parameters for a given combination of $V$ and $\tilde{g}$. $a_\mathrm{x, min}\le0$ and $a_\mathrm{y, max}\ge0$ define the corners of the diamond-shaped form and are the maximum achievable absolute values of $\tilde{a}_\mathrm{x}$ and $\tilde{a}_\mathrm{y}$, respectively. $a_\mathrm{x, max}>0$ is an upper limit for the longitudinal acceleration and is predominantly influenced by the drag force. A fourth parameter $1\le p\le2$ determines the form of the gg-diagram and morphs between a rhombus for $p=1$ and a circle for $p=2$. In total, a combined apparent acceleration $\tilde{a}_\mathrm{x}$ and $\tilde{a}_\mathrm{y}$ is feasible if it satisfies the following inequalities.
\begin{subequations}
	\label{eq:acc_constr}
	\begin{alignat}{2}
		\tilde{a}_\mathrm{x} &\le a_\mathrm{x, max}(V, \tilde{g})\qquad\qquad\qquad\qquad\qquad\qquad\qquad\label{eq:gg_ax_max}\\
		|\tilde{a}_\mathrm{y}| &\le a_\mathrm{y, max}(V, \tilde{g})\label{eq:gg_ay_max}\\
		|\tilde{a}_\mathrm{x}| &\le |a_\mathrm{x, min}(V, \tilde{g})| \left[1 - \left(\frac{|\tilde{a}_\mathrm{y}|}{a_\mathrm{y,max}(V, \tilde{g})}\right)^{p(V, \tilde{g})}\right]^{\frac{1}{p(V, \tilde{g})}}\hspace{-0.2cm}\text{.}\hspace{-1.0cm}\label{eq:gg_combined}
	\end{alignat}
\end{subequations}
The shapes generated by the inequalities \eqref{eq:gg_ax_max}--\eqref{eq:gg_combined} are visualized in Figure~\ref{fig:gg_diagram} by the opaque lines. For a given combination of $V$ and $\tilde{g}$, the four parameters are optimized with an \ac{NLP} to maximize the covered area while always staying within the true shape to ensure feasibility. Figure~\ref{fig:gg_diagram} shows that the underapproximation almost matches the true shape for higher velocities but fails to reach the high positive accelerations at lower velocities. This mismatch becomes noticeable on tracks with sharp turns that require the vehicle to slow down significantly before and accelerate after the turns. This performance loss is analyzed in Section~\ref{sec:3d_and_gg}.

The accelerations in the state vector \eqref{eq:state_vector} are expressed in the velocity frame $\mathcal{V}$. However, the gg-diagrams provide the admissible apparent accelerations and are therefore not directly applicable. The relation between these quantities is given in \eqref{eq:apparent_acc} with the height of the \ac{CoM} $h$. A complete derivation of \eqref{eq:apparent_acc} is given in the appendix.
\begin{equation}
	\label{eq:apparent_acc}
	\begin{bmatrix}
		\tilde{a}_\mathrm{x} \\
		\tilde{a}_\mathrm{y} \\
		\tilde{g} \\
	\end{bmatrix} = \begin{bmatrix}
		\hat{a}_\mathrm{x} + \dot{\hat{\omega}}_\mathrm{y}h - \hat{\omega}_\mathrm{x}\hat{\omega}_\mathrm{z}h + g\left(c_\mu s_\varphi s_{\hat{\chi}}-s_\mu c_{\hat{\chi}}\right)\\
		\hat{a}_\mathrm{y} + \dot{\hat{\omega}}_\mathrm{x}h + \hat{\omega}_\mathrm{y}\hat{\omega}_\mathrm{z}h + g\left(s_\mu s_{\hat{\chi}} + c_\mu s_\varphi c_{\hat{\chi}}\right)\\
		\dot{w} - \hat{\omega}_\mathrm{y}V + (\hat{\omega}^2_\mathrm{x}- \hat{\omega}^2_\mathrm{y})h + g \left(c_\mu c_\varphi\right)
	\end{bmatrix}\text{.}
\end{equation}

\subsection{Time-Optimal \ac{OCP} for Racing}
\label{sec:time_optimal_ocp}
As in previous time-optimal planning approaches \cite{Lovato2021, Lovato2022, Veneri2020, Gundlach2019, Herrmann2021}, we parameterize the \ac{OCP} with the arc length $s$, resulting in the time-optimal cost functional using \eqref{eq:s_dot}:
\begin{equation}
	J(\mathbf{x}) = \int_{s_0}^{s_\mathrm{e}} \frac{1}{\dot{s}}\,ds\text{.}
\end{equation}
This formulation allows specifying a spatial planning horizon $H$ such that ${s_\mathrm{e} = s_0 + H}$. Minimizing $J(\mathbf{x})$ subject to the dynamics and other constraints results in non-smooth solutions with large acceleration gradients that are infeasible due to actuator limits and violate the \ac{QSS} assumption of small changes between two neighboring points on the trajectory. Moreover, a speed limit $V_\mathrm{max}$ lower than the current vehicle velocity and a corresponding adjustment of the velocity constraint would bring the problem into an infeasible region and could cause a solver failure. Two additional cost terms counteract these problems so that the complete formulation of the \ac{OCP} for the local and global racing line planning is
\begin{subequations}
	\label{eq:ocp}
	\begin{alignat}{2}
		&\!\min_{\mathbf{x}, \mathbf{u}}& & \int_{s_0}^{s_\mathrm{e}} \frac{1}{\dot{s}}+\mathbf{u}^\top\mathbf{R}\mathbf{u} +\begin{bmatrix}
			1\\
			\epsilon\\
		\end{bmatrix}^\top\mathbf{S}\begin{bmatrix}
			1\\
			\epsilon\\
		\end{bmatrix}\,ds\label{eq:cost_funciton}\\
		& & &\text{with }\mathbf{R}=\begin{bmatrix}
			w_\mathrm{j,x} & 0 \\
			0 & w_\mathrm{j,y}
		\end{bmatrix}, \mathbf{S}=\begin{bmatrix}
			0 & \frac{w_{\epsilon,1}}{2} \\
			\frac{w_{\epsilon,1}}{2} & w_{\epsilon, 2}
		\end{bmatrix}\nonumber\\
		&\text{s.t.} & & \mathbf{x}^\prime = \frac{d\mathbf{x}}{ds} = \mathbf{f}(\mathbf{x}, \mathbf{u})\frac{1}{\dot{s}}\label{eq:constr_dynamics}\\
		& & & V - \epsilon\le V_\mathrm{max}\text{ with }\epsilon \ge 0\text{ (only local racing line)}\label{eq:constr_V}\\
		& & & \text{\eqref{eq:gg_ax_max}, \eqref{eq:gg_ay_max}, \eqref{eq:gg_combined} with \eqref{eq:apparent_acc}}\label{eq:constr_acc}\\
		& & & n_\mathrm{r}(s) + d_\mathrm{s} \le n \le n_\mathrm{l}(s) - d_\mathrm{s}\label{eq:constr_bound}\\
		& & & -\frac{\pi}{2} \le \hat{\chi} \le \frac{\pi}{2}\label{eq:constr_chi}\\
		& & & \mathbf{x}(s_0=0) = \mathbf{x}(s_\mathrm{e}=s_\mathrm{f}) \text{ (only global racing line)}\text{.}\hspace{-1.0cm}\label{eq:constr_global}
	\end{alignat}
\end{subequations}
The weights $w_\mathrm{j,x}$ and $w_\mathrm{j,y}$ in the second cost term regularize the jerk and prevent the large acceleration gradients. In combination with constraint \eqref{eq:constr_V}, the third cost term corresponds to a soft constraint on the velocity to keep the problem solvable in the case of ${V_\mathrm{max}<V(s_0)}$. In the affected planning steps, the slack variable $\epsilon$ is initialized with the difference $V(s_0) - V_\mathrm{max}$ and otherwise with $0$. It enters the cost function with the linear and quadratic contributions $w_{\epsilon, 1}\epsilon$ and $w_{\epsilon, 2}\epsilon^2$ to reduce the velocity excess within a planning step. The combination of a linear and quadratic term causes a solution with $V\le V_\mathrm{max}$ and $\epsilon=0$ to be preferred if such a solution exists without violating the other constraints. The principles of this approach are shown in \cite{Kerrigan2000} for quadratic cost functions and linear systems. Although it is not guaranteed to hold for our nonlinear cost function and system dynamics, it yields satisfactory results.

The equality constraint \eqref{eq:constr_dynamics} forces the solution to follow the dynamics of the point mass model in \eqref{eq:dynamics} transformed into the spatial domain. The inequality constraint \eqref{eq:constr_acc} summarizes the acceleration constraints. They are implemented by means of a linear interpolation of $a_\mathrm{x, min}$, $a_\mathrm{x, max}$, $a_\mathrm{y, max}$, and $p$ on a uniform mesh of $V$ and $\tilde{g}$. The constraint \eqref{eq:constr_bound} limits the trajectory between the left and right track boundaries. The safety distance $d_\mathrm{s}$ accounts for possible tracking errors and provides a margin for a subsequent local planning approach that considers other vehicles. Constraint \eqref{eq:constr_chi} prohibits reversed driving as a solution. 

The last constraint \eqref{eq:constr_global} is only active for the global racing line generation to ensure a continuous transition at the start-finish line. In contrast, the velocity constraint \eqref{eq:constr_V} and the third cost term are only used for the generation of the local racing line. Here, $\mathbf{x}(s_0)$ is set to the current state of the vehicle and $\mathbf{x}(s_\mathrm{e})$ with $s_\mathrm{e}=s_0 + H$ is free.

\section{Results and Discussion}
\label{sec:results}
\begin{figure}
	\small
	\centering
	\def\axiswidth{6.0cm}
	\input{figures/track.tikz}
	\caption{\acl{MPCB}}
	\label{fig:track}
\end{figure}
The following analyses are carried out on two tracks. The \acf{MPCB}, shown in Figure~\ref{fig:track}, offers both slope and banking angles up to \SI{10}{\degree}. The second track is the \ac{LVMS}, an oval race track with banking angles up to \SI{20}{\degree} but no slopes. The parameters for generating the gg-diagrams and other parameters regarding the implementation are listed in Table~\ref{tab:parameters} in the appendix. The global racing line generation is implemented with CasADi \cite{Andersson2019} using the interface to IPOPT \cite{Wachter2006}. The local racing line planning approach is implemented with acados \cite{Verschueren2022} using the provided \ac{SQP} method.\footnote{A Python implementation of the local and global racing line generation algorithms, the gg-diagram generation, and the track smoothing can be found at \url{https://github.com/TUMRT/online_3D_racing_line_planning}.} The following specifications regarding the computing times are based on simulations with an AMD Ryzen 7 PRO 4750U.

Section~\ref{sec:negligable_terms} identifies negligible terms to reduce the computation time during the online application. The effects of \ac{3D} track geometries and the ability of the proposed gg-diagram underapproximations to represent the true shapes are analyzed in Section~\ref{sec:3d_and_gg}. Section~\ref{sec:online_replanning} presents exemplary results of the online generated local racing lines. Finally, Section~\ref{sec:speed_limit} shows the ability to handle speed limits lower than the current speed.

\subsection{Negligible Terms}
\label{sec:negligable_terms}
\begin{table}
	\caption{Negligible terms based on lap time and constraint violation}
	\setlength\tabcolsep{3pt}
	\begin{center}
		\begin{tabular}{|c|c|c|c|c|c|}
			\cline{3-6}
			\multicolumn{2}{c|}{} & \multicolumn{2}{c|}{\textbf{Lap time}} & \multicolumn{2}{c|}{\textbf{Violation of}}\\
			\multicolumn{2}{c|}{} & \multicolumn{2}{c|}{\textbf{in \si{\second}}} & \multicolumn{2}{c|}{\textbf{\eqref{eq:gg_combined} in \si{\meter\per\second\squared}}}\\
			\hline
			\textbf{\#} & \textbf{Neglected terms} & \ac{MPCB} & \ac{LVMS} & \ac{MPCB} & \ac{LVMS}\\
			\hline
			1 & None & $119.893$ & $27.116$ & $0.00$ & $0.00$\\
			2& $w\hat{\omega}_\mathrm{x}$, $-w\hat{\omega}_\mathrm{y}$ in \eqref{eq:dynamics} & $119.940$ & $27.116$ & $0.00$ & $0.00$ \\
			3& $\dot{\hat{\omega}}_\mathrm{x}h$, $\dot{\hat{\omega}}_\mathrm{y}h$ in \eqref{eq:apparent_acc} & $119.862$ & $27.116$ & $0.78$ & $0.12$ \\
			4& $\hat{\omega}_\mathrm{x}\hat{\omega}_\mathrm{z}h$, $\hat{\omega}_\mathrm{y}\hat{\omega}_\mathrm{z}h$ in \eqref{eq:apparent_acc} & $119.913$ & $27.116$ & $0.22$ & $0.0$ \\
			5& $\hat{\omega}_\mathrm{y}V$ in \eqref{eq:apparent_acc} & $123.181$ & $29.512$ & $12.55$ & $1.40$ \\
			6& $\dot{w}$ in \eqref{eq:apparent_acc} & $120.169$ & $27.126$ & $3.55$ & $0.02$ \\
			7& \#2, \#3, and \#4 & $119.913$ & $27.116$ & $0.77$ & $0.11$\\
			\hline
		\end{tabular}
		\label{tab:negligable_terms}
	\end{center}
\end{table}

As shown in \cite{Lovato2021}, some terms in \eqref{eq:dynamics} and \eqref{eq:apparent_acc} have a non-significant influence on the solution of the \ac{OCP} \eqref{eq:ocp}. Neglecting these terms can reduce the computation time, which is particularly beneficial for the online re-planning scheme. We employ an ablation study with the global racing line generation to identify the terms whose neglect has no or only a minor negative impact on the lap times and constraint violations. Table~\ref{tab:negligable_terms} lists the resulting lap times together with the maximum violation of constraint \eqref{eq:gg_combined} evaluated with all terms. Other constraints are not violated.

The terms \#2, \#3, and \#4 have a non-significant influence on the lap time, and the infeasibility caused by their neglect could be countered by introducing a small margin for the gg-diagrams. On the other hand, term \#5 strongly affects the lap times and compliance with the true acceleration constraints on both tracks. $\hat{\omega}_\mathrm{y}V$ accounts for varying downforces due to banked turns and at apexes of hills and dips. Its magnitude increases with speed and should be taken into account on the given tracks. Finally, term \#6 has a negligible effect on the lap times, but its neglect leads to a significant constraint violation on the \ac{MPCB}. The rapidly changing banking angles due to alternating left and right turns and the additional slope variations on the \ac{MPCB} cause the non-negligible changes in the vertical velocity $w$. On the \ac{LVMS}, however, the vertical velocities and accelerations are small due to the slowly changing banking angles at the turn entries and exits.

The results indicate that the negligible terms are track-dependent. Regarding the computation time of the online planning approach with $N=150$ discretization points, the setup \#7 leads to an average computation time of \SI{80}{\milli\second} with peaks below \SI{100}{\milli\second}. This is in contrast to \#1, with an average computation time of \SI{90}{\milli\second} and peaks above \SI{100}{\milli\second}. If term \#6 was neglected for the \ac{LVMS} as well, an average computation time of \SI{65}{\milli\second} could be achieved. However, we use setting \#7 for both tracks in the following to preserve comparability.

\subsection{\ac{3D} Effects and gg-Diagram Representation}
\label{sec:3d_and_gg}
\begin{figure}
	\small
	\centering
	\def\axiswidth{7.3cm}
	\def\axisheight{3.0cm}
	\input{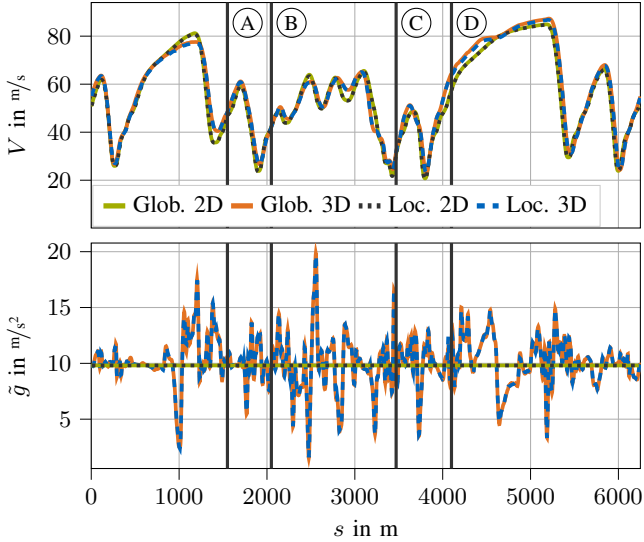}
	\caption{Comparison of $V$ and $\tilde{g}$ profiles for the \ac{MPCB}: Neglecting \ac{3D} effects leads to a different racing line. Following the local (Loc.) racing line that is updated online reproduces the global (Glob.) racing line.}
	\label{fig:rl_profiles}
\end{figure}

First, we analyze how the \ac{3D} track geometry and second how the proposed gg-diagram representation affects the racing line performance. Therefore, we generate the virtual \ac{2D} global racing lines by setting ${\mu(s)=\phi(s)=p_\mathrm{z}(s)=0}$ for ${s\in [0, s_\mathrm{f}]}$, resulting in the lap times corresponding to \#2 in Table~\ref{tab:modi}. On both tracks, neglecting the \ac{3D} geometry results in a considerably larger lap times than the baseline \#1. The velocities and apparent vertical accelerations of the \ac{3D} and \ac{2D} global racing lines on the \ac{MPCB} are visualized in Figure~\ref{fig:rl_profiles}. The velocity of the \ac{2D} global racing line is generally lower in the turns due to the lack of banking. Here, $\tilde{g}$ does not reach a sufficient value to take the turns with larger velocities (e.g., turn \circled{T1} with \SI{9}{\degree} banking). Furthermore, the acceleration of the \ac{3D} global racing line around position \circled{D} cannot be reached without the descent in this section. On the other hand, hill apexes like at $s=\SI{1000}{\meter}$ shortly before \circled{T1} cause a drop in $\tilde{g}$ so that the \ac{3D} racing line cannot accelerate as much as the \ac{2D} racing line. Overall, the \ac{3D} geometries lead to shorter lap times on both tracks.

Regarding the gg-diagram representation, one can observe a performance loss on the \ac{MPCB}. The exact polar coordinate parameterization of the gg-diagrams (\#3) results in a lap time that is almost \SI{6}{\second} shorter than the baseline using our diamond-shaped representation (\#1). This is due to the discrepancy in the available longitudinal accelerations for lower velocities, as mentioned earlier and visualized in Section~\ref{sec:constraints}. However, when used in the online planning approach, the polar coordinate parameterization comes at the cost of required \ac{SQP} iterations and many steps with computation times greater than \SI{100}{\milli\second}. Furthermore, there is no significant performance loss on the oval \ac{LVMS} since the velocity is generally higher and always of a magnitude where the diamond-shaped forms can accurately approximate the true gg-diagrams.

\subsection{Online Replanning}
\label{sec:online_replanning}
\begin{table}
	\caption{Lap times for different global and local racing line generation settings}
	\setlength\tabcolsep{6pt}
	\begin{center}
		\begin{tabular}{|c|c|c|c|}
			\cline{3-4}
			\multicolumn{2}{c|}{} & \multicolumn{2}{c|}{\textbf{Lap time in \si{\second}}} \\
			\hline
			\textbf{\#} & \textbf{Modus} & \ac{MPCB} & \ac{LVMS} \\
			\hline
			1 & Global (\#7 in Table~\ref{tab:negligable_terms}) & $119.913$ & $27.116$ \\
			2 & Global (\acs{2D}) & $124.265$ & $31.876$ \\
			3 & Global (polar) & $114.014$ & $27.124$ \\
			4 & Local & $119.920$ & $27.328$ \\
			5 & Local (\acs{2D}) & $124.661$ & $32.529$ \\
			\hline
		\end{tabular}
		\label{tab:modi}
	\end{center}
\end{table}

The \ac{OCP} for generating the local racing line requires a fixed spatial horizon. The horizon should be as small as possible to achieve low computation times but as large as necessary so that a feasible follow-up trajectory can be found in every planning step. In our case, the emperically determined planning horizons are $H=\SI{300}{\meter}$ and $H=\SI{500}{\meter}$ for the \ac{MPCB} and \ac{LVMS}, respectively. This is in accordance with the common planning horizons used in autonomous racing \cite{Stahl2019,Gundlach2019,Herrmann2021}. A beneficial influence of longer planning horizons on the lap times could not be identified for the considered tracks. All of the following results and lap times are from a flying lap.

First, we verify whether the online approach can achieve the performance of the offline-calculated global racing line. Therefore we assume that the online-generated local racing line can be tracked perfectly until a new local racing line is available. The results \#4 in Table~\ref{tab:modi} show that the online planning approach can nearly reproduce the lap times of the global racing lines on both tracks. Furthermore, the tracked velocities and apparent verticals acceleration match the ones of the global racing lines, as shown in Figure~\ref{fig:rl_profiles} for the \ac{MPCB}.

In a second simulation, we assume an initial state that deviates from the global racing lines' path. Such a scenario is visualized in Figure~\ref{fig:track_obstacles} and can occur during an overtaking maneuver initiated by a local planning approach. Conventional local planning approaches for racing, such as \cite{Rowold2022, Ogretmen2022}, try to return to the global racing line as soon as the local environment allows. However, the time-optimal solution changes due to the new state, so a re-planning of the racing line is required. The solution for the local racing line in the given scenario is visualized in Figures \ref{fig:track_obstacles} and \ref{fig:profiles_obstacles} (Loc. 1). Instead of returning to the global racing line immediately, the time-optimal behavior is to stay on the left side and enter turn \circled{T2} with a slightly lower velocity. Since the planning horizon of \SI{300}{\meter} does not fully cover turn \circled{T3} yet, the end state of the local racing line heads towards the right track boundary and it has a higher velocity compared to the global racing line. Since this undesirable behavior occurs at the end of the planning horizon, it can be corrected in the subsequent planning steps as the horizon moves forward. In the planning step \SI{1}{\second} later (Loc. 2), the horizon covers turn \circled{T3} completely, so that the resulting path and velocity profile return to the global racing line.

\begin{figure}
	\small
	\centering
	\def\axiswidth{8.5cm}
	\input{figures/track_obstacles.tikz}
	\caption{Track section \circled{A} -- \circled{B}: An obstacle on the global racing line (Glob.) causes a lateral offset. The local racing line (Loc.) stays inside when entering turn \circled{T2} since a merge to the outside is worse in terms of time-optimality.}
	\label{fig:track_obstacles}
	\vspace{0.5cm}
	\centering
	\def\axiswidth{7.1cm}
	\def\axisheight{3.1cm}
	\input{figures/profiles_obstacle.tikz}
	\caption{Velocity profiles for the scenario in Figure~\ref{fig:track_obstacles}: The velocity profile of the local racing line adapts with the moving horizon.}
	\label{fig:profiles_obstacles}
\end{figure}

\subsection{Speed Limits}
\label{sec:speed_limit}
\begin{figure}
	\small
	\centering
	\def\axiswidth{8.5cm}
	\input{figures/track_speed_limit.tikz}
	\caption{Track section \circled{C} -- \circled{D}: A speed limit of \SI{20}{\meter\per\second} becomes effective, beginning at the marked position leading to a different behavior.}
	\label{fig:track_speed_limit}
	\vspace{0.5cm}
	\centering
	\def\axiswidth{7.1cm}
	\def\axisheight{3.1cm}
	\input{figures/profiles_speed_limit.tikz}
	\caption{Velocity profiles for the scenario in Figure~\ref{fig:track_speed_limit}: The slack variable $\epsilon$ takes the value of $V-V_\mathrm{max}$ until the speed limit of \SI{20}{\meter\per\second} is reached.}
	\label{fig:profiles_speed_limit}
\end{figure}

To demonstrate the ability to handle speed limits lower than the current velocity, we expose the proposed planning approach to a sudden speed limit of $V_\mathrm{max}=\SI{20}{\meter\per\second}$ during an acceleration phase. Figure~\ref{fig:track_speed_limit} shows the position on the \ac{MPCB} between \circled{C} and \circled{D} from which the speed limit is effective. Reaching this position, the variable $\epsilon$ changes from $0$ to the difference $V-V_\mathrm{max}$, as shown in Figure~\ref{fig:profiles_speed_limit}. It is minimized due to its contribution to the cost function until the velocity excess has been eliminated. During this process, none of the other constraints \eqref{eq:constr_dynamics}, \eqref{eq:constr_acc}--\eqref{eq:constr_chi} is violated. While the online planned trajectory without a speed limit matches the global racing line, a deviation occurs if the velocity is reduced according to $V_\mathrm{max}$. The velocity constraint \eqref{eq:constr_V} and the track boundary constraint \eqref{eq:constr_bound} become the only active constraints so that the shortest path becomes the time-optimal solution, as shown in Figure~\ref{fig:track_speed_limit}.

\section{Conclusion and Outlook}
We present an online \ac{3D} planning approach that updates the racing line based on the current vehicle state and the speed limit. It solves an \ac{OCP} with a \ac{QSS} assumption for a moving horizon. We demonstrate that the approach can reproduce the offline-generated \ac{3D} global racing line during single vehicle racing and that it provides an updated racing line in the case of a deviation from the global racing line. Furthermore, the approach can handle speed limits and adapts the local racing line accordingly for time-optimality under rule compliance.

To make the approach executable online, we identify negligible terms and present an efficient gg-diagram representation to achieve acceptable computation times. The negligible terms strongly depend on the track, and the proposed gg-diagram representation is best suited for oval race tracks with constantly high speeds. To exploit the full acceleration potential on curvy tracks, the gg-diagrams need to approximate the true longitudinal accelerations at lower speeds better.

In future work, we will compare the performance of the local racing line to the global racing line when used as a reference for a subsequent local planning approach that can avoid static obstacles and overtake other vehicles. Scenes like the one visualized in Figure~\ref{fig:track_obstacles} are of interest, for which a quantitative measure could be the remaining lap time. However, this requires a local planning approach that is able to follow the racing lines sufficiently well to reduce the dependency of the results on the local planning approach itself. An alternative approach is to consider obstacles and other vehicles directly in the time-optimal \ac{OCP}.

\bibliographystyle{IEEEtran}
\bibliography{references}

\appendix
The velocity of the \ac{CoM} expressed in the velocity frame is
\begin{equation}
	{}_\mathcal{V}\mathbf{v}_\mathrm{CoM} = \begin{bmatrix}
		V \\ 0 \\ w
	\end{bmatrix} + {}_\mathcal{V}\boldsymbol{\omega}_\mathcal{V} \times \begin{bmatrix}
		0 \\ 0 \\ h
	\end{bmatrix} = \begin{bmatrix}
		V + \hat{\omega}_\mathrm{y}h \\
		\hat{\omega}_\mathrm{x}h \\
		w
	\end{bmatrix}\text{.}
\end{equation}
The acceleration of the \ac{CoM} expressed in the velocity frame is
\begin{equation}
	\label{eq:acc_com}
	\begin{split}
		{}_\mathcal{V}\mathbf{a}_\mathrm{CoM} &= {}_\mathcal{V}\dot{\mathbf{v}}_\mathrm{CoM} + {}_\mathcal{V}\boldsymbol{\omega}_\mathcal{V} \times {}_\mathcal{V}\mathbf{v}_\mathrm{CoM}\\
		&=\begin{bmatrix}
			\dot{V} + \dot{\hat{\omega}}_\mathrm{y}h + \hat{\omega}_\mathrm{y}w - \hat{\omega}_\mathrm{x}\hat{\omega}_\mathrm{z}h \\
			\dot{\hat{\omega}}_\mathrm{x}h + \hat{\omega}_\mathrm{z}V+ \hat{\omega}_\mathrm{z}\hat{\omega}_\mathrm{y}h - \hat{\omega}_\mathrm{x}w \\
			\dot{w} + \hat{\omega}^2_\mathrm{x}h - \hat{\omega}_\mathrm{y}V - \hat{\omega}^2_\mathrm{y}h
		\end{bmatrix} \\
		&=\begin{bmatrix}
			\hat{a}_\mathrm{x} + \dot{\hat{\omega}}_\mathrm{y}h - \hat{\omega}_\mathrm{x}\hat{\omega}_\mathrm{z}h \\
			\hat{a}_\mathrm{y} + \dot{\hat{\omega}}_\mathrm{x}h + \hat{\omega}_\mathrm{z}\hat{\omega}_\mathrm{y}h \\
			\dot{w} + (\hat{\omega}^2_\mathrm{x}- \hat{\omega}^2_\mathrm{y})h - \hat{\omega}_\mathrm{y}V 
		\end{bmatrix}\text{.}
	\end{split}
\end{equation}
The gravitational acceleration expressed in the velocity frame is:
\begin{equation}
	\label{eq:g_com}
	{}_\mathcal{V}\mathbf{g} = g\underbrace{\begin{bmatrix}
			-s_\mu c_{\hat{\chi}} + c_\mu s_\varphi s_{\hat{\chi}} \\
			s_\mu s_{\hat{\chi}} + c_\mu s_\varphi c_{\hat{\chi}} \\
			c_\mu c_\varphi
	\end{bmatrix}}_{\mathclap{\text{last column of }\\ \mathbf{R}^\top_\mathrm{z}(\hat{\chi})\mathbf{R}^\top_\mathrm{x}(\varphi)\mathbf{R}^\top_\mathrm{y}(\mu)\mathbf{R}^\top_\mathrm{z}(\theta)}}\text{.} 
\end{equation}
The sum of \eqref{eq:acc_com} and \eqref{eq:g_com} yields \eqref{eq:apparent_acc}.

\begin{table}[h]
	\caption{Vehicle, tire, and optimization parameters}
	\begin{center}
		\begin{tabular}{|c|c|c|c|c|}
			\cline{1-2}\cline{4-5}
			\textbf{Parameter} & \textbf{Value} & & \textbf{Parameter} & \textbf{Value} \\
			\cline{1-2}\cline{4-5}
			m & $\SI{750}{\kilogram}$ & & $C_{\mathrm{D}_\mathrm{A}}$ & $0.725$ \\
			\cline{1-2}\cline{4-5}
			$C_{\mathrm{Lf}_\mathrm{A}}$ & $0.522$ & & $C_{\mathrm{Lr}_\mathrm{A}}$ & $1.034$ \\
			\cline{1-2}\cline{4-5}
			h & $\SI{0.275}{\meter}$ & & w & $\SI{2.971}{\meter}$ \\
			\cline{1-2}\cline{4-5}
			a & $\SI{1.724}{\meter}$ & & b & $\SI{1.247}{\meter}$ \\
			\cline{1-2}\cline{4-5}
			T & $\SI{1.5815}{\meter}$ & & $\epsilon$ & $0.5$ \\
			\cline{1-2}\cline{4-5}
			$\gamma$ & $1.0$ & & $\mathrm{P}_\mathrm{max}$ & $\SI{357}{\kW}$ \\
			\cline{1-2}\cline{4-5}
			$\delta_\mathrm{max}$ & $\SI{24.6}{\degree}$ & & $\mathrm{p}_{\mathrm{Cx}_1}$ & $2.0$ \\
			\cline{1-2}\cline{4-5}
			$\mathrm{p}_{\mathrm{Dx}_1}$ & $1.7168$ & & $\mathrm{p}_{\mathrm{Dx}_2}$ & $-0.289$ \\
			\cline{1-2}\cline{4-5}
			$\mathrm{p}_{\mathrm{Ex}_1}$ & $0.6975$ & & $\mathrm{p}_{\mathrm{Kx}_1}$ & $63.75$ \\
			\cline{1-2}\cline{4-5}
			$\mathrm{p}_{\mathrm{Kx}_3}$ & $0.2891$ & & $\lambda_{\mu_x}$ & $0.93$ \\
			\cline{1-2}\cline{4-5}
			$\mathrm{p}_{\mathrm{Cy}_1}$ & $1.603$ & & $\mathrm{p}_{\mathrm{Dy}_1}$ & $1.654$ \\
			\cline{1-2}\cline{4-5} 
			$\mathrm{p}_{\mathrm{Dy}_2}$ & $-0.1783$ & & $\mathrm{p}_{\mathrm{Ey}_1}$ & $-1.409$ \\
			\cline{1-2}\cline{4-5} 
			$\mathrm{p}_{\mathrm{Ky}_1}$ & $-53.05$ & & $\mathrm{p}_{\mathrm{Ky}_2}$ & $4.1265$ \\
			\cline{1-2}\cline{4-5} 
			$\lambda_{\mu_y}$ & $0.84$ & & $w_{\mathrm{j,x}}$ & $0.01$\\
			\cline{1-2}\cline{4-5} 
			$w_\mathrm{j,y}$ & $0.01$ & & $w_{\epsilon,1}$ & $60.0$\\
			\cline{1-2}\cline{4-5} 
			$w_{\epsilon,2}$ & $6.0$ & & $d_\mathrm{s}$ & $\SI{0.5}{\meter}$\\
			\cline{1-2}\cline{4-5} 
			$H$ (\ac{MPCB}) & $\SI{300}{\meter}$ & & $H$ (\ac{LVMS}) & $\SI{500}{\meter}$\\
			\cline{1-2}\cline{4-5}
		\end{tabular}
		\label{tab:parameters}
	\end{center}
\end{table}
The meaning of the vehicle and tire parameters in Table~\ref{tab:parameters} modeling the Dallara AV-21 in Figure~\ref{fig:banked_turn} can be found in \cite{Veneri2020}.

\end{document}